\documentclass[11pt,letterpaper]{article}
\usepackage{emnlp2017}
\usepackage{times}
\usepackage{latexsym}

\usepackage{natbib}

\usepackage{tikz}
\usetikzlibrary{external}
\usepackage{pgfplots}
\usetikzlibrary{patterns}
\usetikzlibrary{matrix}
\usetikzlibrary{positioning}

\usepackage{hyperref}

\emnlpfinalcopy
\def\emnlppaperid{13}

\title{Low-Resource Neural Headline Generation}

\author{Ottokar Tilk \and Tanel Alum{\"a}e \\
Department of Software Science, School of Information Technologies,\\Tallinn University of Technology, Estonia \\
  {\tt ottokar.tilk@ttu.ee, tanel.alumae@ttu.ee}}

\date{}

\begin{document}

\maketitle

\begin{abstract}
 Recent neural headline generation models have shown great results, but are generally trained on very large datasets.
 We focus our efforts on improving headline quality on smaller datasets by the means of pre-training.
 We propose new methods that enable pre-training all the parameters of the model and utilize all available text,
 resulting in improvements by up to 32.4\% relative in perplexity and 2.84 points in ROUGE.
 
\end{abstract}

\section{Introduction}


Neural headline generation (NHG) is the process of automatically generating a headline based on the text of the document using artificial neural networks.

Headline generation is a subtask of text summarization. While a summary may cover multiple documents, generally uses similar style to the summarized document, and consists of multiple sentences, headline, in contrast, covers a single document, is often written in a different style (Headlinese \cite{maardh1980headlinese}), and is much shorter (frequently limited to a single sentence).

Due to shortness and specific style, condensing the the document into a headline often requires the ability to paraphrase which makes this task a good fit for abstractive summarization approaches where neural networks based attentive encoder-decoder \cite{bahdanau2014attention} type of models have recently shown impressive results (e.g., \citet{rush2015neural, nallapati2016abstractive}).

While state-of-the art results have been obtained by training NHG models on large datasets like Gigaword, access to such resources is often not possible, especially when it comes to low-resource languages. In this work we focus on maximizing performance on smaller datasets with different pre-training methods.

One of the reasons to expect pre-training to be an effective way to improve performance on small datasets, is that NHG models are generally trained to generate headlines based on just a few first sentences of the documents \cite{rush2015neural, shen2016neural, chopra2016abstractive, nallapati2016abstractive}. This leaves the rest of the text unutilized, which can be alleviated by pre-training subsets of the model on full documents. Additionally, the decoder component of NHG models can be regarded as a language model (LM) whose predictions are biased by the external information from the encoder. As a LM it sees only headlines during training, which is a small fraction of text compared to the documents. Supplementing the training data of the decoder with documents via pre-training might enable it to learn more about words and language structure.

Although, some of the previous work has used pre-training before \cite{nallapati2016abstractive,alifimoffabstractive}, it is not fully explored how much pre-training helps and what is the optimal way to do it. Another problem is, that in previous work only a subset of parameters (usually just embeddings) is pre-trained leaving the rest of the parameters randomly initialized.

The main contributions of this paper are:
LM pre-training for fully initializing the encoder and decoder (sections \ref{sec:encoder_pretraining} and \ref{sec:decoder_pretraining});
combining LM pre-training with distant supervision \cite{mintz2009distant} pre-training using filtered sentences of the documents as noisy targets (i.e. predicting one sentence given the rest) to maximally utilize the entire available dataset and pre-train all the paramters of the NHG model (section \ref{sec:distant_pretraining});
and analysis of the effect of pre-training different components of the NHG model (section \ref{sec:analysis}).

\section{Method}
The model that we use follows the architecture described by \citet{bahdanau2014attention}. Although originally created for neural machine translation, this architecture has been successfully used for NHG (e.g., by \citet{shen2016neural,nallapati2016abstractive} and in a simplified form by \citet{chopra2016abstractive}).

The NHG model consists of:
a bidirectional \cite{schuster1997bidirectional} encoder with gated recurrent units (GRU) \cite{cho2014gru};
a unidirectional GRU decoder;
and an attention mechanism and a decoder initialization layer that connect the encoder and decoder \cite{bahdanau2014attention}.

\begin{figure}
\centering  
\begin{tikzpicture}
\tikzstyle{layer_enc_emb} = [draw, dashed, fill=gray!50, shape=rectangle, align=center, rounded corners, minimum height=2em];
\tikzstyle{layer_enc} = [draw, fill=gray!50, shape=rectangle, align=center, rounded corners, minimum height=2em];
\tikzstyle{layer_dec_emb} = [draw, dashed, pattern=dots, pattern color=gray, shape=rectangle, align=center, rounded corners, minimum height=2em];
\tikzstyle{layer_dec} = [draw,pattern=dots, pattern color=gray, shape=rectangle, align=center, rounded corners, minimum height=2em];
\tikzstyle{layer} = [draw, shape=rectangle, align=center, rounded corners, minimum height=2em];

\matrix (encoder_inputs) [row sep=0em, column sep=0em]
{
    \node (x1) [] {$x_1$}; &
    \node (x1N) [] {$\dots$}; &
    \node (xN) [] {$x_N$}; \\
};
\node (encoder_emb) [layer_enc_emb, above of=encoder_inputs] {Enc. emb.};
\node (encoder) [layer_enc, above =1em of encoder_emb] {Encoder};

\node (attention) [layer, right =2em of encoder] {Attention};
\node (init) [layer, below =1em of attention] {Init.};

\matrix (decoder_inputs) [row sep=0em, column sep=0em, right =8em of x1N]
{
    \node (y1) [] {$y_1$}; &
    \node (y1tm1) [] {$\dots$}; &
    \node (ytm1) [] {$y_{t-1}$}; \\
};
\node (decoder_emb) [layer_dec_emb, above of=decoder_inputs] {Dec. emb.};
\node (decoder) [layer_dec, above =1em of decoder_emb] {Decoder};
\node (yt) [above of=decoder] {$y_t$};

\draw[->, thick] (x1N) -- (encoder_emb);
\draw[->, thick] (encoder_emb) -- (encoder);
\draw[->, thick] (encoder) -- (attention);
\draw[->, thick] (decoder) -- (attention);
\draw[->, thick] (encoder) -- (init);
\draw[->, thick] (init) -- (decoder);
\draw[->, thick] (attention) to[out=90,in=135] (decoder);
\draw[->, thick] (y1tm1) -- (decoder_emb);
\draw[->, thick] (decoder_emb) -- (decoder);
\draw[->, thick] (decoder) -- (yt);

\end{tikzpicture}   
\caption{A high level description of the NHG model. The model predicts the next headline word $y_t$ given the words in the document $x_1 \dots x_N$ and already generated headline words $y_1 \dots y_{t-1}$.}
\label{fig:model}
\end{figure}
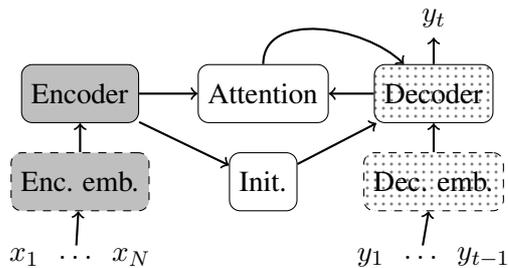

During headline generation, the encoder reads and encodes the words of the document. Initialized by the encoder, the decoder then starts generating the headline one word at a time, attending to relevant parts in the document using the attention mechanism (Figure~\ref{fig:model}). During training the parameters are optimized to maximize the probabilities of reference headlines. 

While generally at the start of training either the parameters of all the components are randomly initialized or only pre-trained embeddings (with dashed outline in Figure~\ref{fig:model}) are used \cite{nallapati2016abstractive,paulus2017deep,gulcehre2016pointing}, we propose pre-training methods for more extensive initialization.

\subsection{Encoder Pre-Training}
\label{sec:encoder_pretraining}
When training a NHG model, most approaches generally use a limited number of first sentences or tokens of the document. For example \citet{rush2015neural, shen2016neural, chopra2016abstractive} use only the first sentence of the document and \citet{nallapati2016abstractive} use up to 2 first sentences. While efficient (training is faster and takes less memory as the input sequences are shorter) and effective (the most informative content tends to be at the beginning of the document \cite{nallapati2016abstractive}), this leaves the rest of the sentences in the document unused. Better understanding of words and their context can be learned if all sentences are used, especially on small training sets.

To utilize the entire training set, we pre-train the encoder on all the sentences of the training set documents.
Since the encoder consists of two recurrent components -- a forward and backward GRU  -- we pre-train them separately. First we add a softmax output layer to the forward GRU and train it on the sentences to predict the next word given the previous ones (i.e. we train it as a LM). After convergence on the validation set sentences, we take the embedding weights of the forward GRU and use them as fixed parameters for the backward GRU. Then we train the backwards GRU following the same procedure as with the forward GRU, with the exception of processing the sentences in a reverse order. When both models are fully trained, we remove the softmax output layers and initialize the encoder of the NHG model with the embeddings and GRU parameters of the trained LMs (highlighted with gray background in Figure~\ref{fig:model}).

\subsection{Decoder Pre-Training}
\label{sec:decoder_pretraining}
Pre-training the decoder as a LM seems natural, since it is essentially a conditional LM.
During NHG model training the decoder is fed only headline words, which is relatively little data compared to the document contents. To improve the quality of the headlines it is essential to have high quality embeddings that are a good semantic representation of the input words and to have a well trained recurrent and output layer to predict sensible words that make up coherent sentences. When it comes to statistical models, the simplest way to improve the quality of the parameters is to train the model on more data, but it also has to be the right kind of data \cite{moore2010intelligent}.

To increase the amount of suitable training data for the decoder we use LM pre-training on filtered sentences of the training set documents. For filtering we use the XenC tool by \citet{Rousseau13XenC} with the cross-entropy difference filtering \cite{moore2010intelligent}. In our case the in-domain data is training set headlines, out-domain data is the sentences from training set documents, and the best cut-off point is evaluated on validation set headlines. The careful selection of sentences is mostly motivated by preventing the pre-trained decoder from deviating too much from Headlinese, but it also reduces training time.

Before pre-training we initialize the input and output embeddings of the LM for words that are common in both encoder and decoder vocabulary with the corresponding pre-trained encoder embeddings.
We train the LM on the selected sentences until perplexity on the validation set headlines stops improving and then use it to initialize the decoder parameters of the NHG model (highlighted with dotted background in Figure~\ref{fig:model}).

A similar approach, without data selection and embedding initialization, has also been used by \citet{alifimoffabstractive}.


\subsection{Distant Supervision Pre-Training}
\label{sec:distant_pretraining}
Approaches described in sections \ref{sec:encoder_pretraining} and \ref{sec:decoder_pretraining} enable full pre-training of the encoder and decoder, but this still leaves the connecting parameters (with white background in Figure~\ref{fig:model}) untrained.

As results in language modelling suggest, surrounding sentences contain useful information to predict words in the current sentence \cite{wang2015larger}. This implies that other sentences contain informative sections that the attention mechanism can learn to attend to and general context that the initialization component can learn to extract.

To utilize this phenomenon, we propose using carefully picked sentences from the documents as pseudo-headlines and pre-train the NHG model to generate these given the rest of sentences in the document. Our pseudo-headline picking strategy consists of choosing sentences that occur within 100 first tokens of the document and were retained during cross-entropy filtering in section \ref{sec:decoder_pretraining}. Picking sentences from the beginning of the document should give us the most informative sentences, and cross-entropy filtering keeps sentences that most closely resemble headlines.

The pre-training procedure starts with initializing the encoder and decoder with LM pre-trained parameters (sections \ref{sec:encoder_pretraining} and \ref{sec:decoder_pretraining}).
After that, we continue training the attention and initialization parameters until perplexity on validation set headlines converges. We then use the trained parameters to initialize all parameters of the NHG model.

Distant supervision has been also used for multi-document summarization by \citet{bravo2012zipf}.


\section{Experiments}
\label{sec:experiments}
We evaluate the proposed pre-training methods in terms of ROUGE and perplexity on two relatively small datasets (English and Estonian).

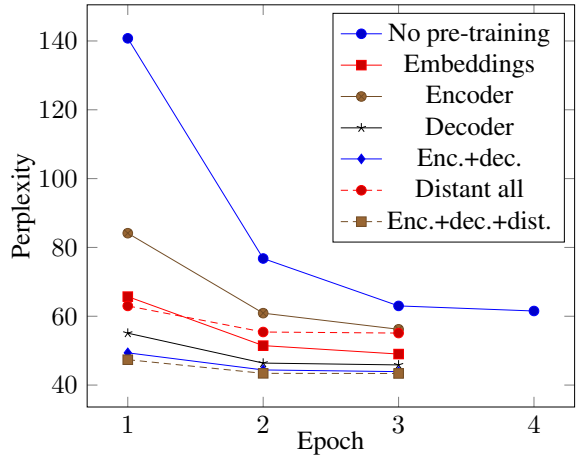
\begin{figure}[t]
\centering  
\resizebox{1.0\linewidth}{!}{
\begin{tikzpicture}
	\begin{axis}[
		xlabel=Epoch,
		ylabel=Perplexity,
		xtick={1,2,3,4,5},
		height=7.5cm,
		ylabel absolute, ylabel style={yshift=-0.3cm},
		xlabel absolute, xlabel style={yshift=0.3cm}
		]

\addplot coordinates {
(1, 140.767304089)
(2, 76.7830234698)
(3, 63.0043404326)
(4, 61.5268161686)
};
\addlegendentry{No pre-training}
\addplot coordinates {
(1, 65.7124126002)
(2, 51.4686774415)
(3, 49.0141545798)
};
\addlegendentry{Embeddings}
\addplot coordinates {
(1, 84.1339625531)
(2, 60.8954038586)
(3, 56.21242778)
};
\addlegendentry{Encoder}
\addplot coordinates {
(1, 55.0854994927)
(2, 46.3873072391)
(3, 45.856500076)
};
\addlegendentry{Decoder}
\addplot coordinates {
(1, 49.369613931)
(2, 44.3965803067)
(3, 43.9038015507)
};
\addlegendentry{Enc.+dec.}
\addplot coordinates {
(1, 63.0011480514)
(2, 55.4177314376)
(3, 55.1123988992)
};
\addlegendentry{Distant all}
\addplot coordinates {
(1, 47.3766163591)
(2, 43.4037550188)
(3, 43.3647987154)
};
\addlegendentry{Enc.+dec.+dist.}
	\end{axis}
\end{tikzpicture}
}
\caption{Validation set (EN) perplexities of the NHG model with different pre-training methods.}
\label{fig:perplexity}
\end{figure}

%
%
%

\begin{table}[t]
\begin{center}
\begin{tabular}{l|ll}
Model & PPL (EN) & PPL (ET) \\ \hline
No pre-training & 65.1 $\pm$1.0 & 25.9 $\pm$0.4 \\\hline
Embeddings & 51.8 $\pm$0.7 & 20.7 $\pm$0.3 \\
Encoder (\ref{sec:encoder_pretraining}) & 59.3 $\pm$0.9 & 23.5 $\pm$0.4 \\
Decoder (\ref{sec:decoder_pretraining}) & 48.3 $\pm$0.7 & 18.8 $\pm$0.3 \\
Enc.+dec. & 46.2 $\pm$0.7 & 17.7 $\pm$0.3 \\
Distant all & 58.6 $\pm$0.9 & 21.3 $\pm$0.3 \\
Enc.+dec.+dist. (\ref{sec:distant_pretraining}) & \textbf{45.8 $\pm$0.7} & \textbf{17.5 $\pm$0.3} \\

\end{tabular}
\end{center}
\caption{Perplexities on the test set with a 95\% confidence interval ~\cite{klakow2002testing}. All pre-trained models are significantly better than the \emph{No pre-training} baseline.}
\label{tab:ppl}
\end{table}

\begin{table*}[t!]
\begin{center}
\begin{tabular}{l|llll|llll}
 & \multicolumn{4}{|c|}{EN} & \multicolumn{4}{c}{ET}\\
Model & $R1_R$ & $R1_P$ & $RL_R$ & $RL_P$ & $R1_R$ & $R1_P$ & $RL_R$ & $RL_P$\\\hline
No pre-training & 20.36 & 33.51 & 17.68 & 29.03 & 26.44 & 34.23 & 25.31 & 32.74\\\hline
Embeddings & \underline{21.09} & 33.36 & 18.23 & 28.72 & \underline{28.42} & \underline{35.94} & \underline{27.02} & \underline{34.16}\\
Encoder (\ref{sec:encoder_pretraining}) & \underline{21.25} & 34.1 & \underline{18.45} & 29.5 & \underline{\textbf{29.28}} & \underline{\textbf{37.04}} & \underline{\textbf{27.88}} & \underline{\textbf{35.24}}\\
Decoder (\ref{sec:decoder_pretraining}) & 20.11 & \underline{31.1} & 17.43 & \underline{26.87} & \underline{25.12} & \underline{32.6} & \underline{23.89} & \underline{30.99}\\
Enc.+dec. & 20.72 & 33.93 & 18.04 & 29.43 & \underline{27.18} & 34.58 & 25.79 & 32.78\\
Distant all & 20.32 & \underline{31.54} & 17.59 & \underline{27.25} & 26.17 & 34.49 & 24.96 & 32.87\\
Enc.+dec.+dist. (\ref{sec:distant_pretraining}) & \underline{\textbf{21.34}} & \underline{\textbf{34.81}} & \underline{\textbf{18.53}} & \underline{\textbf{30.14}} & \underline{27.74} & \underline{35.46} & \underline{26.35} & \underline{33.67}\\

\end{tabular}
\end{center}
\caption{Recall and precision of ROUGE-1 and ROUGE-L on the test sets. Best scores in bold. Results with statistically significant differences (95\% confidence) compared to \emph{No pre-training} underlined.}
\label{tab:rouge}
\end{table*}

\subsection{Training Details}
All our models use hidden layer sizes of 256 and the weights are initialized according to \citet{glorot2010understanding}.
The vocabularies consist of up to 50000 most frequent training set words that occur at least 3 times.
The model is implemented in Theano \cite{bergstra+al:2010-scipy, Bastien-Theano-2012} and trained on GPUs using mini-batches of size 128.
During training the weights are updated with Adam \cite{kingma2014adam} (parameters: $\alpha$=0.001, $\beta_1$=0.9, $\beta_2$=0.999, $\epsilon$=$10^{-8}$ and $\lambda$=$1-10^{-8}$) and $L_2$-norm of the gradient is kept within a threshold of 5.0 \cite{pascanu2013difficulty}.
During headline generation we use beam search with beam size 5.

\subsection{Datasets}
We use the CNN/Daily Mail dataset \cite{nips15_hermann}\footnote{\tt \url{http://cs.nyu.edu/~kcho/DMQA/}} for experiments on English (EN). The number of headline-document pairs is 287227, 13368 and 11490 in training, validation and test set correspondingly.
The preprocessing consists of tokenization, lowercasing, replacing numeric characters with $\#$, and removing irrelevant parts (editor notes, timestamps etc.) from the beginning of the document with heuristic rules.

For Estonian (ET) experiments we use a similarly sized (341607, 18979 and 18977 training, validation and test split) dataset that also consist of news from two sources. During preprocessing, compound words are split, words are truecased and numbers are written out as words. We used Estnltk \cite{estnltkORASMAA16.332} stemmer for ROUGE evaluations.

\subsection{Results and Analysis}
\label{sec:analysis}

\begin{table*}[t!]
\begin{center}
\begin{tabular}{|l|p{0.74\linewidth}|}

\hline
Document & a democratic congressman is at the head of a group of representatives trying to help undocumented immigrants avoid deportations with what they have called the family defender toolkit . the informational pamphlet includes a bilingual card - that some are calling a get out of deportation free card - that lists reasons a person should not be deported under expanded . \\
Reference headline & congressman is developing a get out of deportation toolkit to help undocumented immigrants if they are detained \\
No pre-training & congressman calls for undocumented immigrants \\
Embeddings & congressman calls for help from immigrants trying to help immigrants avoiding deportation \\
Encoder (\ref{sec:encoder_pretraining}) & republican congressman calls for immigrants trying to avoid deportation \\
Decoder (\ref{sec:decoder_pretraining}) & congressman who tried to stop deportations of immigrants \\
Enc.+Dec. & immigration congressman at the head of the head of the group who tries to avoid deportation \\
Distant all & congressman calls for deportation to immigrants who stay in the country \\
Enc.+dec.+dist. (\ref{sec:distant_pretraining}) & congressman tries to help undocumented immigrants avoid deportation \\
\hline

Document & a chihuahua and a bearded dragon showed off their interspecies friendship when they embarked upon a game of tag together . videoed in their front room , the dog named foxxy cleopatra and the reptile called ryuu can be seen chasing after one another around a coffee table . standing perfectly still while looking in the other direction , the bearded dragon initially appears disinterested as the chihuahua jumps around excitedly . \\
Reference headline & you re it! \\
No pre-training & is this the creepiest crawly? \\
Embeddings & meet the poodle! \\
Encoder (\ref{sec:encoder_pretraining}) & it's a knockout! \\
Decoder (\ref{sec:decoder_pretraining}) & the bearded dragon lizard: the bearded dragon lizard spotted in the middle of the street \\
Enc.+Dec. & oh, this is a lion! \\
Distant all & meet the dragon dragon: meet the dragon dragon \\
Enc.+dec.+dist. (\ref{sec:distant_pretraining}) & is this the world's youngest lion? \\
\hline

\end{tabular}
\end{center}
\caption{Examples of generated headlines on CNN/Daily Mail dataset.}
\label{tab:examples}
\end{table*}

Models are evaluated in terms of perplexity (PPL) and full length ROUGE \cite{lin2004rouge}.
In addition to pre-training methods described in sections \ref{sec:encoder_pretraining}-\ref{sec:distant_pretraining}, we also test: initializing only the embeddings using parameters from the LM pre-trained encoder and decoder (\emph{Embeddings}); initializing the encoder and decoder, but leaving connecting parameters randomized (\emph{Enc.+dec.}); pre-training the whole model from random initialization with distant supervision only (\emph{Distant all}); and a baseline that is not pre-trained at all (\emph{No pre-training}).

All pre-training methods gave significant improvements in PPL (Table~\ref{tab:ppl}). The best method (\emph{Enc.+dec.+dist.}) improved the test set PPL by 29.6-32.4\% relative.
Pre-trained NHG models also converged faster during training (Figure~\ref{fig:perplexity}) and most of them beat the final PPL of the baseline already after the first epoch. General trend is that pre-training a larger amount of parameters and the parameters closer to the outputs of the NHG model improves the PPL more. \emph{Distant all} is an exception to that observation as it used much less training data (same as baseline) than other methods.

For ROUGE evaluations, we report ROUGE-1 and ROUGE-L (Table~\ref{tab:rouge}).
In contrast with PPL evaluations, some pre-training methods either don't improve significantly or even worsen ROUGE measures. Another difference compared to PPL evaluations is that for ROUGE, pre-training parameters that reside further from outputs (embeddings and encoder) seems more beneficial. This might imply that a better document representation is more important to stay on topic during beam search while it is less important during PPL evaluation where predicting next target headline word with high confidence is rewarded and the process is aided by previous target headline words that are fed to the decoder as inputs. It is also possible, that a well trained decoder becomes too reliant on expecting correct words as inputs making it sensitive to errors during generation which would somewhat explain why \emph{Enc.+dec.}  performs worse than \emph{Encoder} alone. This hypothesis can be checked in further work by experimenting with methods like scheduled sampling \cite{bengio2015scheduled} that should increase the robustness to mistakes during generation.
Pre-training all parameters on all available text (\emph{Enc.+dec.+dist.}) still gives the best result on English and quite decent results on Estonian. Best models improve ROUGE by 0.85-2.84 points.

Some examples of the generated headlines on the CNN/Daily Mail dataset are shown in Table~\ref{tab:examples}.

\section{Conclusions}
We proposed three new NHG model pre-training methods that in combination enable utilizing the entire dataset and initializing all parameters of the NHG model. We also evaluated and analyzed pre-training methods and their combinations in terms of perplexity (PPL) and ROUGE.
The results revealed that better PPL doesn't necessarily translate to better ROUGE -- PPL tends to benefit from pre-training parameters that are closer to outputs, but for ROUGE it is generally the opposite. Also, PPL benefited from pre-training more parameters while for ROUGE it was not always the case. Pre-training in general proved to be useful -- our best results improved PPL by 29.6-32.4\% relative and ROUGE measures by 0.85-2.84 points compared to a NHG model without pre-training.

Current work focused on maximally utilizing available headlined corpora. One interesting future direction would be to additionally utilize potentially much more abundant corpora of documents without headlines (also proposed by \citet{shen2016neural}) for pre-training.
Another open question is the relationship between the dataset size and the effect of pre-training.

\section*{Acknowledgments}
We would like to thank NVIDIA for the donated GPU, the anonymous reviewers for their valuable comments, and Kyunghyun Cho for the help with the CNN/Daily Mail dataset.

\bibliography{emnlp2017}
\bibliographystyle{emnlp_natbib}

\end{document}